\documentclass[acmsmall,screen]{acmart}
\usepackage{threeparttable}
\usepackage{multirow}
\usepackage{makecell}
\usepackage{bbding}
\usepackage{balance}
\usepackage{xcolor}
\usepackage{multicol}
\usepackage{acronym}
\usepackage{svg}
\AtBeginDocument{%
  }

\setcopyright{acmcopyright}
\copyrightyear{2018}
\acmYear{2018}
\acmDOI{XXXXXXX.XXXXXXX}

\acmConference[Conference acronym 'XX]{Make sure to enter the correct
  conference title from your rights confirmation emai}{June 03--05,
  2018}{Woodstock, NY}
\acmPrice{15.00}
\acmISBN{978-1-4503-XXXX-X/18/06}

\begin{document}

\title[MCT-HFR]{Modality-Collaborative Transformer with Hybrid Feature Reconstruction for Robust Emotion Recognition}
\author{Chengxin Chen}
\email{chenchengxin@hccl.ioa.ac.cn}
\orcid{0000-0003-4510-3313}
\affiliation{%
  \institution{Key Laboratory of Speech Acoustics and Content Understanding, Institute of Acoustics, Chinese Academy of Sciences}
  \streetaddress{No. 21 North 4th Ring Road, Haidian District}
  \city{Beijing}
  \country{China}
  \postcode{100190}
}
\additionalaffiliation{%
  \institution{University of Chinese Academy of Sciences}
  \streetaddress{No. 19(A) Yuquan Road, Shijingshan District}
  \city{Beijing}
  \country{China}
  \postcode{100049}
}

\author{Pengyuan Zhang}
\email{zhangpengyuan@hccl.ioa.ac.cn}
\orcid{0000-0001-6838-5160}
\affiliation{%
  \institution{Key Laboratory of Speech Acoustics and Content Understanding, Institute of Acoustics, Chinese Academy of Sciences}
  \streetaddress{No. 21 North 4th Ring Road, Haidian District}
  \city{Beijing}
  \country{China}
  \postcode{100190}
}
\additionalaffiliation{%
  \institution{University of Chinese Academy of Sciences}
  \streetaddress{No. 19(A) Yuquan Road, Shijingshan District}
  \city{Beijing}
  \country{China}
  \postcode{100049}
}

\renewcommand{\shortauthors}{Chen et al.}

\begin{abstract}
As a vital aspect of affective computing, Multimodal Emotion Recognition has been an active research area in the multimedia community.
Despite recent progress, this field still confronts two major challenges in real-world applications:
1) improving the efficiency of constructing joint representations from unaligned multimodal features, and
2) relieving the performance decline caused by random modality feature missing.
In this paper, we propose a unified framework, Modality-Collaborative Transformer with Hybrid Feature Reconstruction (MCT-HFR), to address these issues. 
The crucial component of MCT is a novel attention-based encoder which concurrently extracts and dynamically balances the intra- and inter-modality relations for all associated modalities.
With additional modality-wise parameter sharing, a more compact representation can be encoded with less time and space complexity. 
To improve the robustness of MCT, we further introduce HFR which consists of two modules: Local Feature Imagination (LFI) and Global Feature Alignment (GFA). 
During model training, LFI leverages complete features as supervisory signals to recover local missing features, while GFA is designed to reduce the global semantic gap between pairwise complete and incomplete representations.
Experimental evaluations on two popular benchmark datasets demonstrate that our proposed method consistently outperforms advanced baselines in both complete and incomplete data scenarios.
\end{abstract}



\received{20 February 2007}
\received[revised]{12 March 2009}
\received[accepted]{5 June 2009}

\maketitle

\section{Introduction}
\ac{MER} is aimed at recognizing and comprehending human emotional states from video clips by leveraging data primarily from three channels: audio (vocal expressions/prosody), vision (facial expressions/gestures), and text (spoken words). 
With the unprecedented advances of social media in recent years, MER has gained increasing attention owing to various promising applications, such as user preference analysis~\cite{Polignano2021app-recommend}, mental health care~\cite{Ayata2020app-medical}, and more empathetic chatting machine~\cite{Zhou2018app-chat}. 
Existing works have explored numerous approaches to fully utilize the complementary information across available modalities in a relatively simple and static multimedia environment~\cite{Abdu2021review}.
However, the real-world scenarios are far more complicated, posing great challenges to MER.

On the one hand, the multimodal features are naturally unaligned due to the varying sampling rates of involved modalities.
One paradigm for handling unaligned multimodal sequences is to first summarize each modality into a high-level representation by collapsing the time dimension.
Afterwards, the cross-modal interactions are modeled using different fusion mechanisms, such as tensor-based operation~\cite{Zadeh2017tfn}\cite{Liu2018lmf}, factorized subspace learning~\cite{Tsai2019mfm}\cite{Hazarika2020misa}, or deep canonical correlation analysis~\cite{Sun2020iccn}\cite{Zhang2021merdcca}.
Albeit, these methods fail to capture the temporal multimodal dynamics. 
For example, a frown and some upset words may appear at the same time, while a sigh may appear at the end of the utterance. 
To this end, several works~\cite{ChenW2017wfrl}\cite{Zadeh2018mfn}\cite{Zadeh2018marn}\cite{Liang2018rmfn} propose to integrate the temporal and cross-modal interactions at each time step of Recurrent Neural Networks (RNNs).
The limitation of these methods is the requirement to \emph{explicitly} align multimodal signals to the equivalent sampling rate in advance, adding extra manual labeling costs.
Inspired by the power of Transformer~\cite{Vaswani2017attention}, researchers have looked into ways to \emph{implicitly} align multimodal sequences during model training~\cite{Tsai2019mult}\cite{Yuan2021tfrnet}\cite{Lv2021pmr}\cite{Guo2022chfn}.
Most of these models are built upon the combination of independent self-attention and cross-attention reinforcement units introduced in~\cite{Tsai2019mult}.
However, fusing multimodal sequences in this directional pairwise manner is inefficient due to limited modality interactions and duplicated model parameters.

On the other hand, the multimodal signals are frequently contaminated in real-world applications.
For instance, the voice may be lost due to the speaker's silence or background noise, the text may be unavailable due to automatic speech recognition errors, and the faces may be invisible due to visual blur or occlusion.
These phenomenons are collectively referred to as \ac{RMFM} in this paper. 
When the RMFM problem arises, existing MER models that are trained on ideal samples typically experience a severe performance decline.
To address this issue, 
researchers have explored different strategies to recover the semantics of missing modalities, 
including missing modality imputation~\cite{Ma2021smil}\cite{Han0KP22mmalign}, generative methods~\cite{Tran2017mmi}\cite{Du2018ssdg}\cite{Zhang2022dpml}, and translation-based methods~\cite{Pham2019mctn}\cite{Zhao2021mmin}\cite{Zeng2022tate}.
The main idea of these approaches is to discover the correlations across modalities, then try to impute the information of missing modalities on the basis of available modalities.
Despite the advancements, these techniques mainly focus on situations where one or two modalities are entirely absent.
In practice, modality missing problem usually occurs at part of the time steps in multimodal sequential signals.
In recent years, 
several approaches~\cite{LiangLTZSM19t2fn}\cite{LiLDZZ20tpfn}\cite{Yuan2021tfrnet}\cite{sun2023emt} have been proposed to handle the fine-grained modality feature missing problem.
However, these works are centered around the scenario where the probabilities of feature missing in training and testing data are comparable.
As a result, different models need to be built for varying feature missing rates.

Motivated by the above observations, we propose \ac{MCT-HFR}, a unified framework for efficient and robust MER in real-world scenarios with uncertain feature missing rates.
We first design the \ac{MCT} for efficient representation learning from unaligned multimodal input.
The key innovation of MCT is the \ac{MRAU}, which takes multiple feature sequences as input and directly returns the reinforced counterparts. 
All modalities in MRAU are dynamically re-scaled with respect to sequence lengths and arranged into a hyper-modality to acquire modality-collaborative information.
The hyper-modality is utilized to reinforce each modality via cross-modal attention. 
Moreover, MRAU can be stacked multiple times to deepen the interactions across modalities.
Compared with the predominant Transformer-based models, MCT avoids the inefficient directional pairwise attention modeling and the repeated parameters of feature projection for the same modality.
On the basis of MCT, we introduce the \ac{HFR} to further strengthen the model robustness towards incomplete data scenarios.
During model training, we apply a dynamic temporal masking on the complete sequences to simulate the RMFM cases, and utilize HFR to reconstruct the missing features.
Specifically, HFR consists of two modules:
1) the \ac{LFI} is designed to reproduce the local missing elements of the input sequences leveraging the reinforced modality features, while 
2) the \ac{GFA} is designed to align the the global semantics summarized from the complete and incomplete views of multimodal input.
Consequently, HFR tries to capture the feature missing patterns from two diverse perspectives, which can further complement each other.
The superiority of our proposed framework is verified by extensive experiments on different multimodal emotion recognition benchmarks, IEMOCAP~\cite{iemocap} and MSP-IMPROV~\cite{mspimprov}, with both complete and incomplete testing data.

The main contributions of this paper can be summarized as follows:
\begin{itemize}
 \item We propose MCT to promote the efficiency of representation learning from unaligned multimodal sequences. With lower time and space complexity, MCT can achieve better performance over existing Transformer-based models.
 \item We propose HFR to strengthen the robustness of MCT in real-world scenarios with uncertain feature missing rates. HFR is intended to reconstruct the missing semantics from both the local and global perspectives to stabilize the representation learning from incomplete multimodal sequences.
 \item Experiments on two benchmark datasets demonstrate that our proposed MCT-HFR outperforms several state-of-the-art models in both complete and incomplete data scenarios, highlighting the efficacy of MCT-HFR for practical applications.
\end{itemize}

The remainder of this paper is organized as follows: 
Section~\ref{related works} reviews the previous related work. 
Section~\ref{preliminary} formalizes the problem statement and introduces the vanilla attention units.
Section~\ref{methodology} describes the overall workflow and detailed components of our proposed method.
Section~\ref{experimental setup} introduces the experimental datasets and setup in detail.
Section~\ref{experimental results} presents the experimental results and analysis. 
Finally, Section~\ref{conclusion} concludes this work and discusses the future research directions.

\section{Related Works} \label{related works}
\subsection{Multimodal Emotion Recognition}
Since humans can express their emotions in a variety of ways, including through tone of voice, facial expressions, and verbal language, it is expected that combining data from several modalities will be more effective than unimodal approaches~\cite{Poria2017review}.
The multimodal fusion approaches can be broadly categorized into three levels: 
1) feature-level fusion, 2) model-level fusion, and 3) decision-level fusion.
Generally, feature-level fusion concatenates the time-aligned features at the front end to build up the mutlimodal input~\cite{RozgicASKP12early}\cite{NojavanasghariG16early}\cite{Georgiou2019early}, while decision-level fusion performs inference based on each modality and subsequently combines the predictions from all modalities~\cite{ZadehZPM16late}~\cite{GuoZLW18late}~\cite{Perez-RuaVPBJ19late}.
Nevertheless, these two types of fusion methods fail to explicitly model cross-modal interactions.
Therefore, various modern fusion strategies have been centered around model-level fusion in the past few years.
One line of research utilized independent encoders to extract high-level representations of different modalities prior to feature fusion~\cite{Zadeh2017tfn}~\cite{Liu2018lmf}~\cite{Hazarika2020misa}~\cite{Tsai2019mfm}~\cite{Sun2020iccn}\cite{Zhang2021merdcca}.
For instance, Hazarika~\emph{et al.}~\cite{Hazarika2020misa} factorize modalities into complementary subspaces to construct a comprehensive multimodal representation.
To further capture fine-grained interactions across modalities, a variety of approaches have been proposed, such as multi-view RNN networks~\cite{ChenW2017wfrl}\cite{Zadeh2018mfn}\cite{Zadeh2018marn}\cite{Liang2018rmfn} and multimodal word embeddings~\cite{Wang2019wordshift}~\cite{Rahman2020mag}~\cite{TsengNG21multimodalembedding}.
Despite the advancements, the requirement for time-aligned features is a bottleneck for these methods in practical use.
More recently, Transformer-based methods~\cite{Tsai2019mult}~\cite{HuangTLLN20multer}\cite{Yuan2021tfrnet}\cite{Lv2021pmr}\cite{Guo2022chfn}~\cite{ChenXXYP22keysparse} have gained increasing popularity because of their capacity to fuse unaligned multimodal features directly.
For example, Tsai~\emph{et al.}~\cite{Tsai2019mult} propose the modality reinforcement unit to repeatedly reinforce a target modality with information from a source modality by learning the directional pairwise attention.
Lv~\emph{et al.}~\cite{Lv2021pmr} introduce a message hub to exchange information with each modality, then iteratively update the common message and the involved modality features.
Based on the message hub, Sun~\emph{et al.}~\cite{sun2023emt} further reduced the time complexity via local-global partial attention mechanism.
Compared with existing approaches 
that build upon the combination of vanilla attention units, our proposed MCT optimizes the attention mechanism specifically for multimodal input, reinforcing the associated modalities collaboratively in a single layer with less time and space complexity.

\subsection{Methods for Random Modality Feature Missing Problem}
While enormous approaches have been proposed to improve MER in the context of ideal data, the generalization of MER in real-world scenarios has yet to be investigated.
Based on the assumption for data imperfection, previous works can be generally categorized into two groups: 1) modality missing, and 2) modality feature missing.
The former one considers the situations where one or two modalities are entirely absent in the testing phase.
In recent years, several approaches have been explored to handle this problem~\cite{Georgiou2021is}\cite{Ma2021smil}\cite{Han0KP22mmalign}\cite{Zhang2022dpml}~\cite{Pham2019mctn}\cite{Zhao2021mmin}\cite{Zeng2022tate}.
For instance, Pham~\emph{et al.}~\cite{Pham2019mctn} propose a robust representation learning strategy leveraging cyclic translations from source to target modalities. 
Moreover, Zeng~\emph{et al.}~\cite{Zeng2022tate} incorporate a tag encoding module with common space projection to address the issue of uncertain missing modalities.
In contrast, modality feature missing considers a more general situation where partial time steps of sequential features are missing.
Under this assumption, Liang~\emph{et al.}~\cite{LiangLTZSM19t2fn} develop a low-rank regularization based model for time-aligned multimodal sequences, and Li~\emph{et al.}~\cite{LiLDZZ20tpfn} further improve the model by exploiting the temporal-wise dynamics of data with less space complexity.
More recently, Yuan~\emph{et al.}~\cite{Yuan2021tfrnet} propose a Transformer-based model with sequential feature reconstruction for unaligned multimodal sequences, and 
Sun~\emph{et al.}~\cite{sun2023emt} integrate the former low-level feature reconstruction with high-level feature attraction to enhance the model robustness. 
Although intriguing, these techniques require distinct models to be trained for various feature missing rates, which limits the models' generalization ability.
In contrast, this work attempts to build one unified framework with superior robustness against \emph{uncertain} feature missing patterns.
In the training phase, the incomplete multimodal features are dynamically generated, 
and the complete view of multimodal input is referenced for the reconstruction of missing features.

\subsection{Self-supervised Representation Learning}
\ac{SSL}, a promising paradigm for representation learning without requiring manually labeled data, has advanced natural language processing~\cite{deepword}\cite{bert}\cite{ClarkLLM20electra}\cite{HeLGC21deberta}, speech processing~\cite{abs-1807-03748cpc}\cite{SchneiderBCA19wav2vec}\cite{BaevskiZMA20wav2vec2}\cite{wavlm}, and computer vision~\cite{ChenK0H20sslcv1}\cite{CaronTMJMBJ21sslcv2}\cite{RadfordKHRGASAM21cv3}\cite{Bao0PW22cv4}.
By setting up the appropriate pretext tasks, SSL methods enable models to learn the inherent structure or patterns present in unlabeled data.
For example, 
masked prediction is designed to learn meaningful contextualized representations by predicting the masked elements of data, \emph{e.g.}, word embeddings~\cite{bert}\cite{ClarkLLM20electra}, quantized speech units~\cite{BaevskiZMA20wav2vec2}\cite{HsuBTLSM21hubert}, and discretized visual tokens~\cite{RadfordKHRGASAM21cv3}\cite{Bao0PW22cv4}.
Contrastive learning is another approach to learn semantically rich yet discriminative representations by bringing similar samples closer and separating diverse samples apart~\cite{ChenK0H20sslcv1}\cite{cvpr/0010KBLY21cl2}\cite{cvpr/YangLZXLYG22cl3}.
Generally, the pre-trained models can be used as seed models and then optimized for downstream tasks.
Moreover, Hendrycks~\emph{et al.}~\cite{HendrycksMKS19sslrobust} reveal that substantial regularization may be achieved by integrating self-supervision and full supervision, which enhances robustness and uncertainty estimation.
Inspired by the power of SSL, our proposed HFR sets up a pretext task as an auxiliary supervision, which reconstructs the missing semantics of incomplete multimodal sequences from both the local and global perspectives. The pretext task is trained jointly with the major classification task.

\section{Preliminary} \label{preliminary}
In this section, we first formalize the problem and its corresponding notations. 
Subsequently, we briefly introduce the vanilla attention units.

\subsection{Problem Definition}
Our task is to recognize the emotion expressed in videos leveraging three major modalities, including audio ($a$), vision ($v$), and language ($l$). 
The complete feature sequences are denoted by $\mathbf{\overline{X}}_m\in \mathbb{R}^{T_m \times d_m}$, where $T_m$ and $d_m$ represent the sequence length and feature dimension of the corresponding modality $m \in \{ a, v, l\}$, respectively.
The feature masking indicators are denoted by $\mathbf{M}_m\in \mathbb{R}^{T_m}$, where 1 denotes ablating the feature vector at this time step, while 0 denotes keeping it unaltered.
The derived incomplete feature sequences are denoted by $\mathbf{X}_m = \mathbf{\overline{X}}_m \odot \mathbf{M}_m \in \mathbb{R}^{T_m \times d_m}$.
Hence, the problem can be formulated as how to map $\left( \mathbf{X}_{a}, \mathbf{X}_{v}, \mathbf{X}_{l} \right)$ into 
the emotion label $\mathbf{y} \in \mathbb{R}^c$ for each video segment.
Additionally, $\left( \mathbf{\overline{X}}_{a}, \mathbf{\overline{X}}_{v}, \mathbf{\overline{X}}_{l} \right)$ and $\left( \mathbf{M}_{a}, \mathbf{M}_{v}, \mathbf{M}_{l} \right)$ are utilized as auxiliary references for the feature reconstruction during model training.

\subsection{Vanilla Attention Units}
Transformer~\cite{Vaswani2017attention} is proposed to directly models the dependence between any pairwise frames of sequential features via the scaled dot-product attention.
In the \ac{SAU}, the input sequences $\mathbf{X}_{t} \in \mathbb{R}^{T_t \times d_t}$ are first transformed into queries, keys, and values, which are denoted by $\mathbf{Q}_{t} = \mathbf{X}_{t}\mathbf{W}_{Q}$ with $\mathbf{W}_{Q} \in  \mathbb{R}^{d_t \times d_k}$, $\mathbf{K}_{t} = \mathbf{X}_{t}\mathbf{W}_{K}$ with $\mathbf{W}_{K} \in  \mathbb{R}^{d_t \times d_k}$, and $\mathbf{V}_{t} = \mathbf{X}_{t}\mathbf{W}_{V}$ with $\mathbf{W}_{V} \in  \mathbb{R}^{d_t \times d_v}$.
Afterwards, the output of SAU is calculated as:
\begin{align}
\mathbf{Y}_{t}
 =\operatorname{SAU}\left( \mathbf{X}_{t} \right) 
 = \operatorname{softmax}\left( \frac{\mathbf{Q}_t \mathbf{K}_t^\top}{\sqrt{d_k}} \right) \mathbf{V}_t
\end{align}
where $\mathbf{Y}_{t} \in \mathbb{R}^{T_t \times d_v}$. 
Further, multi-head attention is introduced in \cite{Vaswani2017attention} to jointly learn diverse relationships between queries and keys from different representation sub-spaces.

The \ac{CAU} is introduced to reinforce the features of target modality from another feature sequences of source modality ~\cite{Tsai2019mult}, denoted as $\mathbf{X}_{t} \in \mathbb{R}^{T_t \times d_t}$ and $\mathbf{X}_{s} \in \mathbb{R}^{T_s \times d_s}$, respectively. 
Different from SAU, the queries in CAU are mapped from the target modality, denoted as $\mathbf{Q}_{t} = \mathbf{X}_{t}\mathbf{W}_{Q}$, while the keys and values are mapped from the source modality, denoted as $\mathbf{K}_{s} = \mathbf{X}_{s}\mathbf{W}_{K}$, and $\mathbf{V}_{s} = \mathbf{X}_{s}\mathbf{W}_{V}$, respectively.
Hence, the output of CAU can be formulated as:
\begin{align}
\mathbf{Y}_{t}
 =\operatorname{CAU}\left( \mathbf{X}_{s}, \mathbf{X}_{t} \right) 
 = \operatorname{softmax}\left( \frac{\mathbf{Q}_t \mathbf{K}_s^\top}{\sqrt{d_k}} \right) \mathbf{V}_s
\end{align}

\section{Methodology} \label{methodology}
\begin{figure*}[htb]
  \centering
  \scalebox{1.0}
  {\includegraphics[width=\linewidth]{ 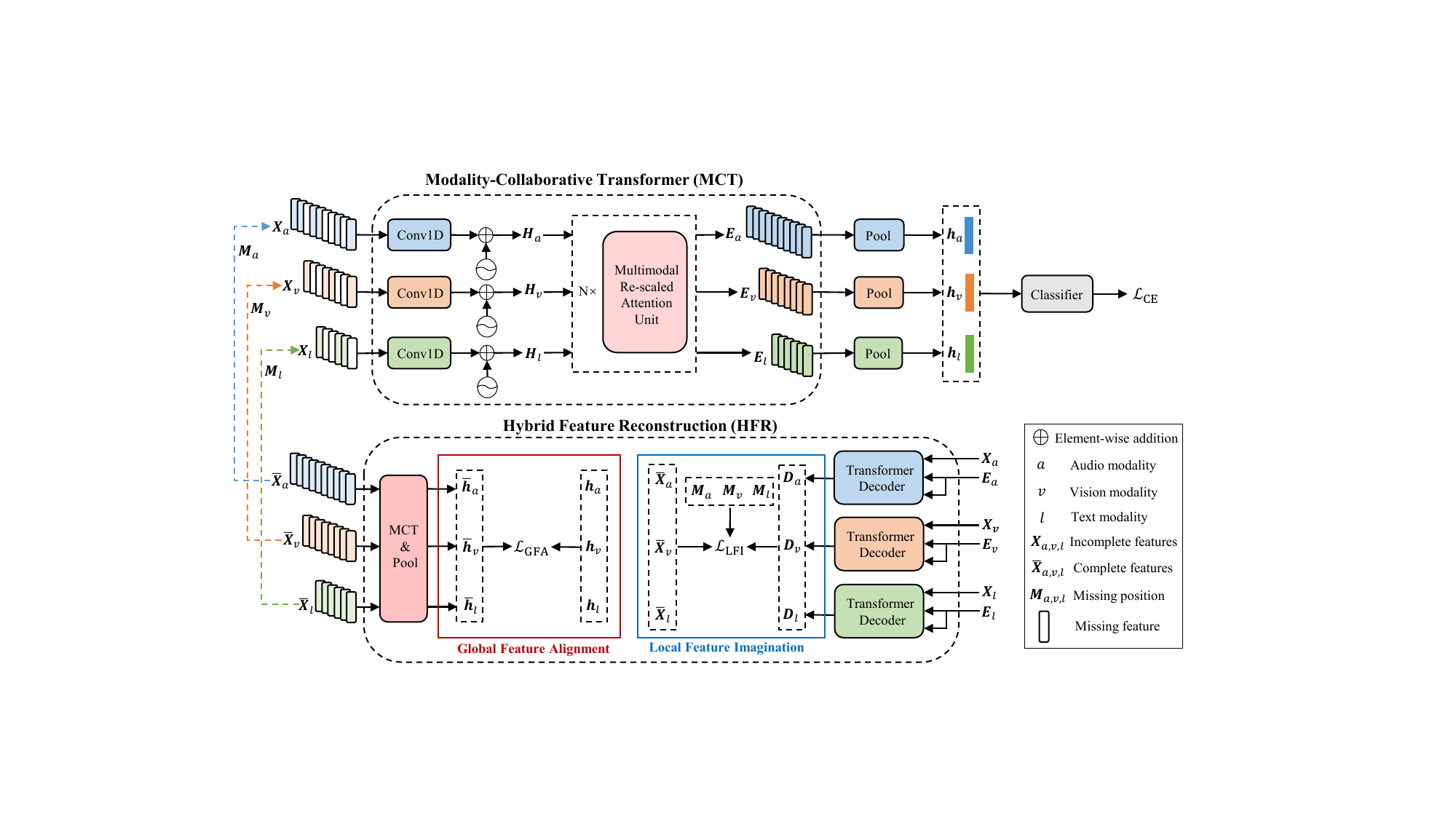}}
  \caption{The overview framework of our proposed Modality-Collaborative Transformer with Hybried Feature Reconstruction.}
  \label{fig:overview}
\end{figure*}

As depicted in Fig.~\ref{fig:overview}, the overall workflow of MCT-HFR consists of two major branches:
1) \emph{Emotion recognition task related branch:}
The \ac{MCT} is utilized to transform the incomplete sequences into reinforced modality representations, which are then summarized into global vectors for the emotion category classification. 
This branch is activated throughout the training and testing phases;
2) \emph{Feature reconstruction related branch:}
In the \ac{LFI} module, the imagined modality representations are generated by corresponding Transformer-based decoders, then we calculate the reconstruction loss of the local missing features.
In the \ac{GFA} module, we also extract the global vectors of complete sequences to compute the similarity loss with the counterparts of incomplete sequences. 
Together, these two modules make up the \ac{HFR}.
This branch is activated only in the training phase.
In the following sub-sections, we present the details of each module.

\subsection{Modality-Collaborative Transformer} \label{MCT}

\subsubsection{Unimodal Encoding Module}
To capture the neighborhood information of the input sequence $X_m$, a 1D convolution ($\operatorname{Conv1D}$) layer is first employed on the temporal dimension:
\begin{align}
\mathbf{X}^{\prime}_m = \operatorname{Conv1D}\left(\mathbf{X}_m, k_m\right) \in \mathbb{R}^{T_m \times d}
\end{align}
where $k_m$ is the convolution kernel size, $m \in \{ a, v, l\}$.
In this way, the input sequences are projected into corresponding semantic spaces with a shared feature dimension $d$.
We further integrate the element-wise position information into the projected features using the sinusoidal position embedding (PE)~\cite{Vaswani2017attention}:
\begin{align}
\mathbf{H}_m = \mathbf{X}^{\prime}_m + \operatorname{PE}\left(T_m, d\right)
\end{align}
where $\operatorname{PE}\left(T_m, d\right) \in \mathbb{R}^{T_m \times d}$ computes the embeddings for each time step.

\begin{figure*}[htb]
  \centering
  \scalebox{0.7}
  {\includegraphics[width=\linewidth]{ 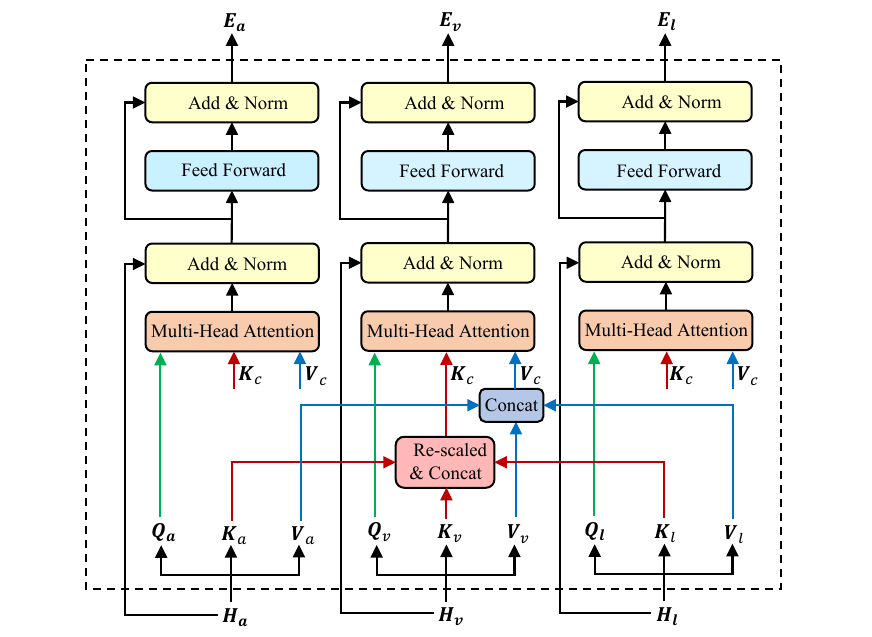}}
  \caption{Detailed diagram of the Multimodal Re-scaled Attention Unit.}
  \label{fig:mrau}
\end{figure*}

\subsubsection{Multimodal Re-scaled Attention Unit}
The architecture of the \ac{MRAU} is shown in Fig.~\ref{fig:mrau}.
In order to efficiently capture the underlying correlations between elements across modalities, we first construct a hyper-modality that integrates the information from all the modalities.
Later, each modality is reinforced by the hyper-modality using cross-modal attention. 
Formally, MRAU takes multiple feature sequences as input and returns the \emph{collaboratively} reinforced features:
\begin{align}
\mathbf{E}_{a}, \mathbf{E}_{v}, \mathbf{E}_{l} = \operatorname{MRAU} \left( \mathbf{H}_{a}, \mathbf{H}_{v}, \mathbf{H}_{l}\right)
\end{align}
where $\mathbf{E}_m \in \mathbb{R}^{T_m \times d}$, $m \in \{ a, v, l\}$.
Note that the input and output sequences in each MRAU layer share the same feature shapes, thus multiple layers could be stacked to deeply reinforces the involved modalities.

For each MRAU layer, the input features are first projected into the queries, keys, and values, respectively:
$\mathbf{Q}_{m} = \mathbf{H}_{m}\mathbf{W}_{Q}^m$, 
$\mathbf{K}_{m} = \mathbf{H}_{m}\mathbf{W}_{K}^m$, 
and $\mathbf{V}_{m} = \mathbf{H}_{m}\mathbf{W}_{V}^m$.
Intuitively, the hyper-modality can be regarded as the source modality of cross-modal attention, 
and the associated keys and values can be obtained via temporal-wise concatenation.
However, the simple concatenation of unaligned sequences may lead to inferior performance. 
For each modality $m$, the feature length $T_m$ is variable with a maximum length of $T_m^{\operatorname{max}}$ during training.
On the one hand, the attention scores are calculated collaboratively across all modalities. 
Due to the varied sampling rate of different modalities, the longer sequence of a certain modality may be assigned with larger accumulated attention weights, degrading the importance of other modalities.
On the other hand, if the testing samples greatly exceed the maximum length, the increased length of concatenated sequence may distract the attention from emotionally salient parts.

To this end, we propose a dynamic re-scaling strategy on the keys of each modality:
1) \emph{Attention balance factor:} 
We first introduce $\gamma_b = \frac{1}{\sqrt{T_m}}$ as a penalty to balance the relative importance of different modalities in computing the attention scores; 
2) \emph{Attention extrapolation factor:} 
In~\cite{chiang-cholak-2022-overcoming}, Chiang et al. proposed a log-length scaled attention to improve the extrapolation ability of self-attention. 
Inspired by this work, we extend the log-length factor into a multimodal form, denoted as $\gamma_e = \log_{\sum_{m}T_m^{\operatorname{max}}} {\textstyle \sum_{m}T_m}$, to overcome the attention dispersion problem.
Hence, the construction of $\mathbf{K}_{c}$ and $\mathbf{V}_{c}$ can be formulated as:
\begin{align}
\mathbf{K}_{m}^{\prime} &= \gamma_b \cdot \gamma_e \cdot \mathbf{K}_{m} \\
\mathbf{K}_{c} &=  \operatorname{Concat} \left(\left[ \mathbf{K}_{a}^{\prime} ;  \mathbf{K}_{v}^{\prime} ; \mathbf{K}_{l}^{\prime}\right]\right) \\
\mathbf{V}_{c} &=  \operatorname{Concat} \left( \left[ \mathbf{V}_{a} ;  \mathbf{V}_{v} ; \mathbf{V}_{l}\right] \right)
\end{align}

Afterwards, the computation in MRAU can be formulated as:
\begin{align}
\mathbf{H}_{m}^{\prime} &= \operatorname{LN} \left( \operatorname{softmax}\left( \frac{\mathbf{Q}_m {\mathbf{K}_c}^\top}{\sqrt{d_k}} \right) \mathbf{V}_c + \mathbf{H}_{m}  \right)  \\
\mathbf{E}_{m} &= \operatorname{LN} \left( \operatorname{FFN} \left(\mathbf{H}_{m}^{\prime} \right)+ \mathbf{H}_{m}^{\prime}\right) 
\end{align}
where LN denotes layer normalization~\cite{BaKH16LN}, and FFN is the position-wise feed-forward network in Transformer.

Note that the projections of $\mathbf{Q}_{m}$, $\mathbf{K}_{m}$, $\mathbf{V}_{m}$ occur at the beginning of a single MRAU layer.
They are calculated only once for each modality and shared for later modality reinforcements in the same MRAU layer.
Compared with the predominant Transformer-based approaches that independently model the interactions between pairs of modalities, MRAU can reduce the modality reinforcement units from ``$2M$ CAU + $2M$ SAU''~\cite{Lv2021pmr}\cite{sun2023emt} to ``$M$ CAU'' for $M$ modalities.
As a result, MRAU benefits from less space and time complexity, especially when more modalities are involved.

\subsection{Pooling and Classification}
Finally, the attention-based pooling~\cite{MirsamadiBZ17attnpool} is applied to summarize the reinforced modality features into global representations.
The derived vectors, denoted as $\mathbf{h}_a$, $\mathbf{h}_v$ and $\mathbf{h}_l$, are concatenated as the overall multimodal representation $\mathbf{h}$ for the downstream emotion classifier:
\begin{align}
\mathbf{y}^{\prime} = \operatorname{softmax} \left(  \operatorname{FC} \left(\mathbf{h}, \theta_c \right)\right)
\end{align}
where $\mathbf{y}^{\prime} \in \mathbb{R}^{C}$ is the estimated probabilities, $C$ is the number of classes, and FC denotes fully connected layers.
Afterwards, the standard Cross-Entropy (CE) loss is employed for the classification task:
\begin{align}
\mathcal{L}_{\operatorname{CE}} = - \sum_{i=1}^{C} y_i\log{ y^{\prime}_i}
\end{align}
where $y_i$ and $y_i^{\prime}$ are the ground-truth and predicted scores for each class $i$.

\subsection{Hybrid Feature Reconstruction} \label{HFR}
\subsubsection{Local Feature Imagination}
Given the incomplete multimodal sequences, the LFI module is aimed at directing MCT in regenerating the missing fine-grained features of associate modalities.
As illustrated in Section~\ref{MCT}, the collaboratively reinforced features $\mathbf{E}_m$ are expected to carry sufficient multimodal contextual information.
Subsequently, we employ the Transformer-based decoders to reconstruct the complete view of each modality:
\begin{align}
\mathbf{D}_m = \operatorname{CAU}\left(\mathbf{E}_m,\operatorname{SAU}(\mathbf{X}_m) \right)
\end{align}
where $\mathbf{D}_m$ denotes the regenerated sequences leveraging original input and reinforced output of MCT.
With the help of collaboratively reinforced features, it’s promising to reproduce the almost ``identical'' local features.
In~\cite{Girshick_2015_ICCV}, Girshick et al. introduced a smooth L1 loss, which is less sensitive to outliers than L1 loss and prevents exploding gradients in some cases. Mathematically, we have:
\begin{align}
\operatorname{smooth}_{\operatorname{L1}}\left( x \right) = \left\{\begin{matrix}
0.5x^2 \quad &\operatorname{if} \left |x\right |<1 \\
\left | x \right |-0.5 &\operatorname{otherwise}
\end{matrix}\right.
\end{align}

Following prior works~\cite{Yuan2021tfrnet}\cite{sun2023emt}, LFI also adopts smooth L1 loss to measure the difference between the imagined missing features and the associated ground-truth features. Hence, the overall loss function of LFI can be formulated as:
\begin{align}
\mathcal{L}_{\operatorname{LFI}} = \textstyle \sum_{m\in\{ a,v,l\}} \operatorname{smooth}_{\operatorname{L1}}\left(\left( \mathbf{\overline{X}}_m-\mathbf{D}_m \right) \cdot \mathbf{M}_m\right)
\end{align}
where the feature masking indicators $\mathbf{M}_m$ are utilized to constraint the loss function at the missing positions.

\subsubsection{Global Feature Alignment}
The final goal of MCT is to enrich the emotional semantics of the global multimodal representations, which is not directly captured by recovering the local semantics of each modality. 
Inspired by the philosophy of contrastive learning, the GFA module is intended to reduce the discrepancy between the latent representations summarized from the complete and incomplete views of data.
Specifically, both the complete and incomplete multimodal features are utilized to compute the global vectors, $\mathbf{h}$ and $\overline{\mathbf{h}}$, respectively.
Afterwards, we adopt the \ac{CMD}~\cite{ZellingerGLNS17CMD} metric,
which is an advanced distance metric that compares the order-wise moment differences of two representations to assess the discrepancy in their distributions.
Mathematically, we have:
\begin{align}
\mathcal{L}_{\operatorname{CMD}}\left( \mathbf{x}_1, \mathbf{x}_2\right) = \left \| \operatorname{E}\left( \mathbf{x}_1\right) - \operatorname{E}\left( \mathbf{x}_2\right) \right \|_{2}+\sum_{k=2}^{K}\left \| \operatorname{C}_k\left( \mathbf{x}_1\right) - \operatorname{C}_k\left( \mathbf{x}_2\right) \right \|_{2} 
\end{align}
where $\operatorname{E}\left( \mathbf{x}\right)$ is the empirical expectation vector of the input sample $\mathbf{x}$, and $\operatorname{C}_k\left( \mathbf{x}\right) =\operatorname{E}\left(\left( \mathbf{x} -\operatorname{E}\left( \mathbf{x}\right)\right)^k\right)$ denotes the vector of $k^{\operatorname{th}}$ order sample central moments with respect to the coordinates.
By optimizing the distance metric, the global semantics of complete and incomplete views are implicitly aligned.
For comparison, we implement another two popular distance metrics in contrastive learning, including cosine similarity and \ac{JSD}. However, they lead to inferior performance (see in Section~\ref{ablation}).

To improve the quality of representations summarized from the complete view, we also calculate the associate CE loss as an additional supervision.
Hence, the overall loss function of GFA can be formulated as follows:
\begin{align}
\mathcal{L}_{\operatorname{GFA}} = -\sum_{i=1}^{C} y_i\log{ \overline{y}^{\prime}_i} + \mathcal{L}_{\operatorname{CMD}}\left( \operatorname{FC} \left(\mathbf{h}, \theta_p \right), \overline{\mathbf{h}}\right) 
\end{align}
where FC is utilized for the semantic space mapping.

\subsection{Model Training}
\subsubsection{Overall Loss Function}
In the proposed MCT-HFR framework, the above representation learning process is guided by three different supervisions.
The classification loss of incomplete data $\mathcal{L}_{\operatorname{CE}}$ is the primary supervision for emotion recognition.
While the feature reconstruction losses (including $\mathcal{L}_{\operatorname{GFA}}$ and $\mathcal{L}_{\operatorname{LFI}}$) on the complete-incomplete data pairs are regarded as the auxiliary supervisions for robust representation learning.
Combining all the objectives together, the overall loss function can be formulated as follows:
\begin{align}
\mathcal{L} = \mathcal{L}_{\operatorname{CE}} + \alpha\mathcal{L}_{\operatorname{GFA}} +\beta\mathcal{L}_{\operatorname{LFI}} 
\end{align}
where $\alpha$ and $\beta \in \mathbb{R}$ are the hyper-parameters that determine the contribution of hybrid feature reconstruction.
\subsubsection{Dynamic Incomplete Training}
Different from prior works~\cite{Yuan2021tfrnet}\cite{sun2023emt}, 
the incomplete training data are dynamically generated on the fly in this work.
The probability of sequential feature masking is denoted as $p_{\operatorname{miss}}$, which follows the Bernoulli distribution.
Besides, the proportion of incomplete data in a training batch increases linearly from 0.0 to 1.0 in the first 5 epochs.
We empirically find that $p_{\operatorname{miss}}=0.2$ will suffice in our evaluation.

\section{Experimental Datasets and Setup} \label{experimental setup}
\subsection{Dataset Description}
IEMOCAP and MSP-IMPROV are two popular benchmark datasets for multimodal emotion recognition. In Table~\ref{tab:dataset}, we present the emotion distribution in each session of the datasets.

\textbf{IEMOCAP:}
The IEMOCAP dataset~\cite{iemocap} contains approximately 12 hours of audio-visual data from 10 actors. The dataset contains 5 sessions and each session is performed by one female and male actor in scripted and improvised scenarios. To be consistent with the previous works, we merge the instances labeled ``excited'' into the ``happy'' class.
Moreover, we remove the special notes in the manual transcriptions, such as ``[LAUGHTER]'', ``[BREATHING]'', and ``[GARBAGE]''.

\textbf{MSP-IMPROV:} 
The MSP-IMPROV dataset~\cite{mspimprov} contains 6 sessions of dyadic interactions between pairs of male-female actors. 
15 target sentences are used to collect the recordings. 
For each target sentence, 4 emotional scenarios were created to elicit happy, angry, sad, and neutral responses. 
We remove the videos shorter than 1 second, and select the videos in the “Other-improvised” group which were recorded in improvised scenarios.
Compared with IEMOCAP, the emotion distribution of MSP-IMPROV is more unbalanced.
Besides, the emotion behaviors presented in MSP-IMPROV are more natural and spontaneous, making it a more challenging dataset.

\begin{table}[htbp]
\centering
\caption{Emotion distribution in each session of the IEMOCAP and MSP-IMPROV datasets.}
\label{tab:dataset}
\begin{tabular}{lccccccccccc}
\toprule
\multirow{2}{*}{Emotion} & \multicolumn{5}{c}{IEMOCAP Session No.} & \multicolumn{6}{c}{MSP-IMPROV Session No.} \\
\cmidrule(lr){2-6} \cmidrule(lr){7-12}
                         & 1      & 2      & 3     & 4     & 5     & 1    & 2     & 3     & 4     & 5    & 6    \\
\midrule                         
Angry (A)                & 229    & 137    & 240   & 327   & 170   & 54   & 54    & 73    & 52    & 119  & 108  \\
Happy (H)                & 278    & 327    & 286   & 303   & 442   & 92   & 162   & 143   & 140   & 238  & 224  \\
Neutral (N)              & 384    & 362    & 320   & 258   & 384   & 204  & 284   & 409   & 169   & 309  & 358  \\
Sad (S)                  & 194    & 197    & 305   & 143   & 245   & 76   & 78    & 73    & 76    & 109  & 215  \\
\midrule
Total                    & 1085   & 1023   & 1151  & 1031  & 1241  & 426  & 578   & 698   & 437   & 775  & 905 \\
\bottomrule
\end{tabular}
\end{table}

\subsection{Feature Extraction}
We first pre-process the raw video segments by extracting the features of each modality for both datasets.
The detailed feature extraction process is shown as follows:
\subsubsection{Text Modality}
Recently, Transformer-based pre-trained language models has gained great success in the field of natural language processing~\cite{sslnlp}.
In this paper, a pre-trained BERT~\cite{bert} model~\footnote{\url{https://huggingface.co/bert-base-uncased}} is utilized to encode the sentences into deep linguistic features.
As suggested by previous works~\cite{deepword}, the average results of hidden states from the last four layers are calculated as the final word-level features $\mathbf{X}_{l}$ with $d_l$ equal to 768.
In order to simulate the transcription errors, the missing tokens are replaced by the unknown token ``[UNK]'' in BERT.

\subsubsection{Audio Modality}
Similar to the  feature extraction process of text modality, a pre-trained wav2vec 2.0~\cite{BaevskiZMA20wav2vec2} model~\footnote{\url{https://huggingface.co/facebook/wav2vec2-base-960h}} is exploited as the acoustic feature extractor. 
The derived acoustic features $\mathbf{X}_{a}$ are sampled at a frame rate of 50 \emph{Hz} with $d_a$ equal to 512. 
Finally, the missing frames are replaced by zero vectors.

\subsubsection{Vision Modality}
To start with, OpenFace toolkit~\cite{Baltrusaiti2018openface} is used to extract aligned faces from the raw videos at a frame rate of 5 \emph{Hz}.
Afterwards, a ResNet50~\cite{He2016resnet} model~\footnote{\url{https://github.com/cydonia999/VGGFace2-pytorch.git}} fine-tuned on VGGFace2~\cite{Cao2018vggface} is utilized to further extract deep visual features.
Global average pooling is applied on the output of the last convolution layer to derive the feature vector of a certain facial expression.
The time steps with no detected faces are padded with zero vectors to obtain the frame-level features $\mathbf{X}_{v}$ with $d_v$ equal to 512.
Finally, the missing frames are replaced by zero vectors.

\subsection{Implementation Details} \label{implementation}
Our proposed model was implemented using the PyTorch~\cite{pytorch} framework on a single Tesla P100 GPU (12 GB).
For parallel computing, the multimodal features of the same mini-batch were truncated or padded to the equal-length, and the batch size was set to 32.
Concretely, the maximum lengths of each modality $\left( T_a^{\operatorname{max}}, T_v^{\operatorname{max}}, T_l^{\operatorname{max}} \right)$ were set to $\left( 400, 40, 50 \right)$, respectively.
The output channels $d$ of $\operatorname{Conv1D}$ were set to 128, and the kernel sizes $\left( k_a, k_v, k_l \right)$ were set to $\left( 3, 3, 1 \right)$.
The stacked MRAU layers ($N$) were set to 4, and the cross-modal attention comprised of 4 heads with 32 nodes $\left( d_k \right)$ in each head.
An AdamW optimizer~\cite{LoshchilovH19AdamW} with a learning rate of $10^{-4}$ was applied to optimize the model parameters.
To determine the weight of different losses, we adjusted the hyper-parameters $\alpha$ and $\beta$ from 0 to 1 with a step length of 0.2.
The models were trained for a maximum of 40 epochs, and the training was stopped if the validation loss did not decrease for 8 consecutive epochs.

Following prior works, leave-one-session-out cross-validation (LOSO CV) was employed on both datasets. In each fold, the data from a certain session was split by speakers into the validation and testing sets, while the rest sessions were used for training.
All models were trained using five different random seeds for a fair comparison, and the average results are reported in this paper.

\subsection{Baseline Models} \label{baseline}
To validate the effectiveness of our proposed method, we reproduce the following state-of-the-art models as the baselines.
All the approaches are applicable on unaligned multimodal sequences.

\textbf{MulT.}
Multimodal Transformer (MulT)~\cite{Tsai2019mult} employs directional pairwise cross-modal attention to repeatedly reinforce a target modality with information from the other modalities. 

\textbf{MISA.}
By projecting each sample's modality into two subspaces, this method learns both Modality-Invariant and -Specific Representations (MISA)~\cite{Hazarika2020misa} which benefit the multimodal fusion.

\textbf{PMR.}
Progressive Modality Reinforcement (PMR)~\cite{Lv2021pmr} introduces a message hub to exchange information with each modality by reinforcing the common messages and modality-specific semantics in an iterative way.

\textbf{CHFN.} 
Cross Hyper-modality Fusion Network (CHFN)~\cite{Guo2022chfn} generates the multimodal-shifted word representations to dynamically capture the variations of nonverbal contexts. 

\textbf{TFR-Net.}
Transformer-based Feature Reconstruction Network (TFR-Net)~\cite{Yuan2021tfrnet} integrates self-attention reinforcement on the basis of MulT as the feature encoder, and introduces an MLP-based temporal feature reconstruction module to tackle the modality feature missing problem.

\textbf{EMT-DLFR.}
Efficient Multimodal Transformer with Dual-Level Feature Restoration (EMT-DLFR)~\cite{sun2023emt} introduces a mutual promotion unit with local-global partial attention mechanism to reduce computational complexity, and integrates the former MLP-based temporal feature reconstruction~\cite{Yuan2021tfrnet} with high-level feature attraction to enhance the model robustness.

The training methods described in the original papers for these baselines can be divided into two categories:
1) ``Conventional complete training'' is employed in MulT, MISA, PMR and CHFN. This strategy trains the models only on the complete data;
2) ``One-to-one training'' is employed in TFR-Net and EMT-DLFR. This strategy generates the incomplete training data statically before model training, and the feature missing rate is aligned with the testing data. 
Hence, different models need to be trained for evaluation as the missing rate of testing data changes.
By contrast, our proposed ``dynamic incomplete training'' only trains one model, and this model is evaluated on the testing data with arbitrary missing rate.

\subsection{Evaluation Metrics}
In this research, the performance of emotion recognition is evaluated using four metrics: unweighted average accuracy (UA), weighted
average accuracy (WA), unweighted average F1 (UF1) and weighted average F1 (WF1). 
Concisely, UA/UF1 is referred to as a
mean of accuracies/F1-scores for various emotion classes, and WA/WF1 is a weighted mean accuracies/F1-scores over various emotion classes with weights proportionate to the number of instances in a class~\cite{prml}. Given the inherent class imbalance present in both the IEMOCAP and MSP-IMPROV datasets, UA, UF1 and WF1 are adopted
as the major evaluation metrics, while WA is considered as a supplementary assessment metric.
For all the metrics, higher values indicate better model performance.

To evaluate the overall performance of model in incomplete data scenarios, we compute the Area Under Indicators Line Chart (AUILC) value for each metric following~\cite{Yuan2021tfrnet}.
Given the model evaluation metric scores $\{ s_0, s_1, \dots, s_t\}$ with increasing missing rates $\{ r_0, r_1, \dots, r_t\}$, AUILC value is defined as:
\begin{align}
\operatorname{AUILC} = \sum_{i=0}^{t-1} \frac{\left( s_{i+1}+s_i\right)}{2}\cdot \left( r_{i+1}-r_i\right).
\end{align}
On both datasets, we evaluate the model under the feature missing rates of $\{0.0, 0.1, \dots, 0.9\}$.

\section{Experimental Results and Analysis} \label{experimental results}

\subsection{Comparison With Existing Works}
\subsubsection{Classification Performance Analysis} 
First of all, we perform the quantitative experiments and evaluate different approaches in both complete and incomplete data scenarios.
Table~\ref{tab:accuracy_iemocap} and Table~\ref{tab:accuracy_msp-improv} present the experimental results on the IEMOCAP and MSP-IMPROV datasets, respectively. 
For fair comparison, these models can be divided into:
traditional MER models using ``conventional complete training'' strategy (G-\uppercase\expandafter{\romannumeral1}),  
MER models designed for modality feature missing using ``one-to-one training'' strategy (G-\uppercase\expandafter{\romannumeral2}), 
and our proposed MCT-HFR with all the three training strategies (G-\uppercase\expandafter{\romannumeral3}).
Note that MCT-HFR is equivalent to MCT when using the ``conventional complete training'' strategy.
From the experimental results, we conclude the following observations:

\begin{table}[htbp]
\centering
\caption{Performance of different approaches on the IEMOCAP dataset using 5-fold LOSO CV. 
For the incomplete testing data, the AUILC values of each metric with missing rate intervals $\{0.0, 0.1, \dots, 0.9\}$ are reported.
Models with $^\star$ were trained using ``dynamic incomplete training'' strategy, while models with $^\dagger$ were trained using ``one-to-one training'' strategy.
The best results are highlighted in bold.}
\label{tab:accuracy_iemocap}
\begin{tabular}{lcccccccc}
\toprule
\multicolumn{1}{c}{\multirow{2}{*}{Models}} & \multicolumn{4}{c}{Scores(\%) on Complete Testing Data} & \multicolumn{4}{c}{Scores(\%) on Incomplete Testing Data}\\
\cmidrule(lr){2-5} \cmidrule(lr){6-9}
\multicolumn{1}{c}{} & UA ($\uparrow$)  & UF1 ($\uparrow$)  & WA ($\uparrow$)  & WF1 ($\uparrow$)  & UA ($\uparrow$)  & UF1 ($\uparrow$)  & WA ($\uparrow$)  & WF1 ($\uparrow$)      \\
\midrule                        
MulT            & 76.39  & 76.05  & 75.88  & 75.83    & 63.63  & 63.69  & 63.50  & 63.46  \\
MISA            & 77.13  & 77.01  & 76.75  & 76.66    & 64.23  & 64.25  & 64.03  & 63.95  \\
PMR            & 77.36  & 77.30  & 77.15  & 77.09    & 65.36  & 65.08  & 65.01  & 64.85 \\
CHFN             & 77.05  & 76.85  & 76.86  & 76.64    & 63.14  & 63.31   & 63.09  & 63.09 \\
\midrule                        
TFR-Net$^\dagger$           & 77.05  & 76.40  & 76.33  & 76.10    & 66.64  & 65.92  & 65.85  & 65.61  \\
EMT-DLFR$^\dagger$           & 77.48  & 77.07  & 77.13  & 76.87    & 67.36  & 66.96  & 66.94  & 66.73  \\
\midrule
MCT  & 78.68 & 77.75  & 77.92  & 77.39    & 66.47  & 65.81  & 65.74  & 65.51 \\
MCT-HFR$^\dagger$                & 78.68 & 77.75  & 77.92  & 77.39    & 67.78  & 67.51  & 67.38  & 67.25 \\
MCT-HFR$^\star$      & \textbf{78.85}  & \textbf{78.43}  & \textbf{78.41}  & \textbf{78.21}    & \textbf{68.06}  & \textbf{67.80}  & \textbf{67.68}  & \textbf{67.57} \\       
\bottomrule
\end{tabular}
\end{table}

\begin{table}[htbp]
\centering
\caption{Performance of different approaches on the MSP-IMPROV dataset using 6-fold LOSO CV. 
For the incomplete testing data, the AUILC values of each metric with missing rate intervals $\{0.0, 0.1, \dots, 0.9\}$ are reported.
Models with $^\star$ were trained using ``dynamic incomplete training'' strategy, while models with $^\dagger$ were trained using ``one-to-one training'' strategy.
The best results are highlighted in bold.}
\label{tab:accuracy_msp-improv}
\begin{tabular}{lcccccccc}
\toprule
\multicolumn{1}{c}{\multirow{2}{*}{Models}} & \multicolumn{4}{c}{Scores(\%) on Complete Testing Data} & \multicolumn{4}{c}{Scores(\%) on Incomplete Testing Data}\\
\cmidrule(lr){2-5} \cmidrule(lr){6-9}
\multicolumn{1}{c}{} & UA ($\uparrow$)  & UF1 ($\uparrow$)  & WA ($\uparrow$)  & WF1 ($\uparrow$)  & UA ($\uparrow$)  & UF1 ($\uparrow$)  & WA ($\uparrow$)  & WF1 ($\uparrow$)      \\
\midrule                        
MulT            & 68.50  & 66.49  & 67.00  & 66.76    & 58.27  & 55.96  & 56.17  & 56.31  \\
MISA            & 70.42  & 68.51  & 68.97  & 68.67    & 59.83  & 57.05  & 57.21  & 56.87  \\
PMR            & 68.44  & 69.17  & 70.89  & 70.82    & 58.74  & 58.77  & 59.25  & 59.22 \\
CHFN             & 67.09  & 67.62  & 68.81  & 68.63    & 55.98  & 57.19   & 58.91  & 58.58 \\
\midrule                        
TFR-Net$^\dagger$           & 69.61  & 68.02  & 68.40  & 68.02    & 61.21  & 59.58  & 59.97  & 59.95  \\
EMT-DLFR$^\dagger$           & 69.46  & 69.01  & 70.19  & 70.30    & 60.72  & 60.17  & 60.77  & 60.77  \\
\midrule  
MCT             & \textbf{71.81}  & 70.34  & 70.56  & 70.34    & 60.70  & 59.96  & 60.43  & 60.41 \\
MCT-HFR$^\dagger$    & \textbf{71.81}  & 70.34  & 70.56  & 70.34    & 61.38  & 60.96  & 61.47  & 61.49 \\
MCT-HFR$^\star$      & 71.55  & \textbf{71.93}  & \textbf{71.25}  & \textbf{71.86}    & \textbf{62.26}  & \textbf{61.60}  & \textbf{62.03}  & \textbf{61.92} \\
\bottomrule
\end{tabular}
\end{table}

1) Compared with the models in G-\uppercase\expandafter{\romannumeral1}, our proposed MCT outperforms currently advanced approaches on both datasets.
For the complete testing data, MCT exceeds the best performing baselines by 1.32\%/1.39\% UA and 0.45\%/1.17\% UF1 on the IEMOCAP/MSP-IMPROV datasets, respectively. 
For the incomplete testing data, a similar conclusion can be drawn.
We attribute these promising results to the collaborative modality reinforcement in MCT, which promotes the efficiency of modeling interactions across modalities and fastens the convergence of representation learning.

\begin{figure*}[htb]
  \centering
  \scalebox{1.0}
  {\includegraphics[width=\linewidth]{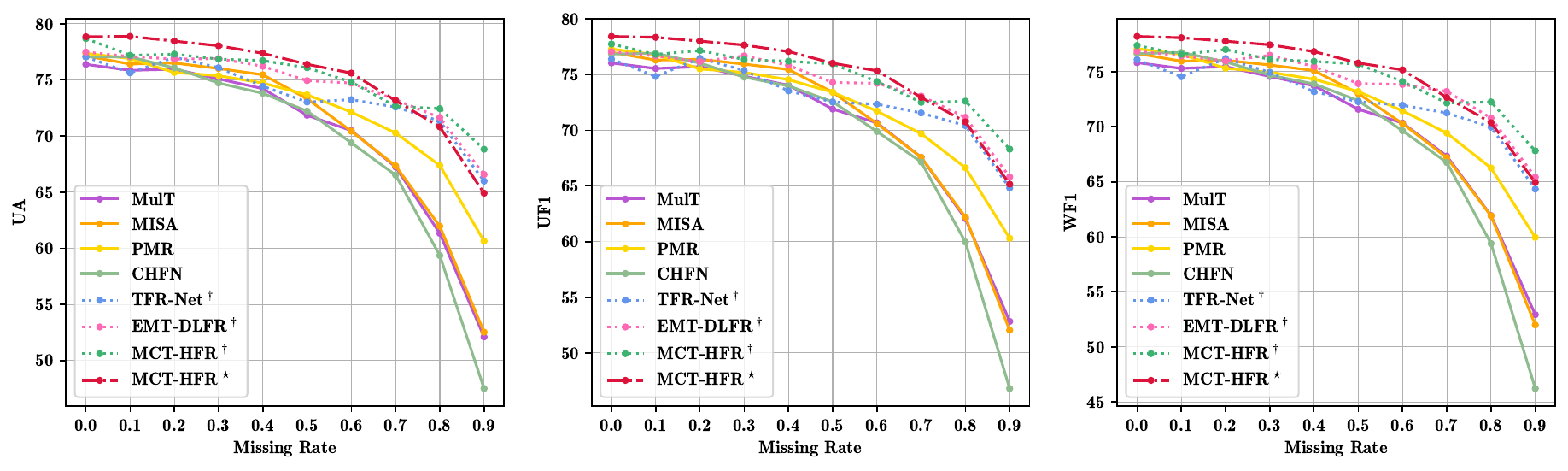}}
  \caption{Performance variation curves of different approaches on the IEMOCAP dataset under different feature missing rates.}
  \label{fig:iemocap}
\end{figure*}

\begin{figure*}[htb]
  \centering
  \scalebox{1.0}
  {\includegraphics[width=\linewidth]{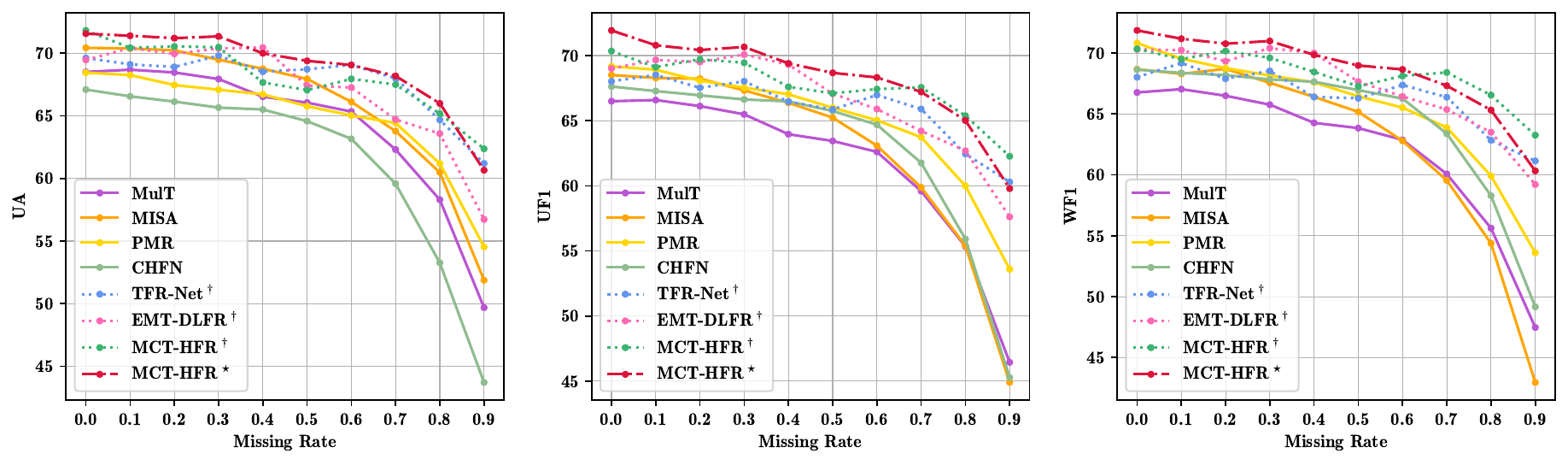}}
  \caption{Performance variation curves of different approaches on the MSP-IMPROV dataset under different feature missing rates.}
  \label{fig:msp-improv}
\end{figure*}

2) Compared with the models in G-\uppercase\expandafter{\romannumeral2}, our proposed MCT-HFR outperforms the state-of-the-art models using the same training strategy on both datasets.
For the incomplete testing data, MCT-HFR surpasses the suboptimal model, EMT-DLFR, by the AUILC values of 0.42\%/0.66\% UA, 0.55\%/0.79\% UF1 and 0.52\%/0.72\% WF1 on the IEMOCAP/MSP-IMPROV datasets, respectively.
Furthermore, we plot the curves of detailed performance with respect to various missing rates in Fig.~\ref{fig:iemocap} and Fig.~\ref{fig:msp-improv}.
As the missing rate increases, the performance decline of our method is smaller than the others.
Taking the results on IEMOCAP for example, as the missing rate increases from 0.0 to 0.9, the performance of baselines decreases by 11.07\%\textasciitilde29.54\% UA and 11.57\%\textasciitilde30.04\% UF1, while MCT-HFR only decreases by 9.86\% UA and 9.44\% UF1, highlighting the robustness of MCT-HFR.
We attribute these encouraging results to both the collaborative modality reinforcement in MCT and the complementary feature reconstruction in HFR.

3) By comparing the models in G-\uppercase\expandafter{\romannumeral3}, we can observe that the introduction of ``dynamic incomplete training'' strategy can further enhances AUILC values for MCT-HFR on the incomplete testing data. 
It is interesting that MCT-HFR$^\star$ can achieve even better performance than MCT-HFR$^\dagger$ for most metrics on the complete testing data.
This phenomenon is highly related to the ``dynamic incomplete training'' strategy.
In some aspects, this strategy is similar to data augmentation, which may alleviate the overfitting problem when the dataset (IEMOCAP and MSP-IMPROV) is relatively small.
Albeit, the performance of MCT-HFR$^\star$ is inferior to MCT-HFR$^\dagger$in severe feature missing cases.
Compared with ``one-to-one training'', the main advantage of ``dynamic incomplete training'' is to present one unified model that covers both complete and most incomplete data scenarios.

\begin{table}[htbp]
\centering
\caption{Computational complexity of different Transformer-based models.
The layer-wise and model-wise parameters and MACs are reported, with the best results highlighted in bold.}
\label{tab:complexity}
\begin{tabular}{lcccccc}
\toprule
\multicolumn{1}{c}{\multirow{2}{*}{Models}} & \multicolumn{2}{c}{Layer-wise} & \multicolumn{2}{c}{Model-wise for Training}& \multicolumn{2}{c}{Model-wise for Inference}\\
\cmidrule(lr){2-3} \cmidrule(lr){4-5} \cmidrule(lr){6-7}
\multicolumn{1}{c}{} & Params (M)  & MACs (G)  & Params (M)  & MACs (G) & Params (M)  & MACs (G)    \\
\midrule                        
MulT            & 0.89 & 0.15 & 7.16  & 1.47 & 7.16 & 1.47  \\
PMR             & 0.99 & 0.32 & \textbf{4.59}  & 1.42 & 4.59 & 1.42  \\
TFR-Net         & 1.34 & 0.22 & 12.25 & 2.67 & 8.95 & 2.14  \\
EMT-DLFR         & 0.99 & 0.08 & 5.26 & \textbf{0.88} & 4.57 & 0.46  \\
\midrule 
MCT-HFR         & \textbf{0.45} & \textbf{0.07} & 4.72  & 1.18 & \textbf{2.41} & \textbf{0.42}  \\                
\bottomrule
\end{tabular}
\end{table}

\subsubsection{Computational Complexity Analysis}
To further investigate the efficiency of our proposed MCT-HFR, we measure the computational complexity via the metrics of trainable parameters and multiply–accumulate operations (MACs)~\footnote{\url{https://pypi.org/project/thop/}}.
We perform the calculations on a random generated multimodal instance with $\left( T_a, T_v, T_l \right)$ equal to $\left( 400, 40, 50 \right)$ and report the results in Table~\ref{tab:complexity}.
For fair comparison, the hyper-parameters of attention units are kept in line (4 layers and 4 attention heads with 32 nodes in each head) across the Transformer-based models.
For TFR-Net, EMT-DLFR, and MCT-HFR, the feature reconstruction networks are omitted during inference.
From the experimental results, we conclude the following observations:

1) We first compare the computational complexity of different fusion networks consisted of one layer. From Table~\ref{tab:complexity}, we find that each layer of MCT-HFR (\emph{i.e.}, an MRAU in Section~\ref{MCT}) enjoys the least trainable parameters (3$\times$ less than TFR-Net and 2$\times$ less than EMT-DLFR) and MACs (3$\times$ less than TFR-Net and slightly less than EMT-DLFR).
Specifically, MulT and TFR-Net are based on directional pairwise attention modeling, requiring $M(M-1)$ and $M^2$ attention-based reinforcement units for $M$ modalities, respectively.
Although PMR reduces the reinforcement units to $2M$, this model constructs a new hyper-modality as the message hub and iteratively updates it along with the original modalities, resulting in the highest MACs.
On the basis of PMR, EMT-DLFR reduced the time complexity via local-global partial attention mechanism.
In contrast, our proposed MCT-HFR further narrows the reinforcement units to $M$, and the projected keys, queries, and values of each modality are shared for later modality reinforcements in the same layer.
Additionally, MCT-HFR dynamically builds a hyper-modality via a parameter-free re-scaling mechanism and regards it as the source modality.
As a result, the trainable parameters and MACs of MCT-HFR can be consistently lowered.

2) Subsequently, we look into the results of overall models (basically including the unimodal encoders, multimodal fusion networks, pooling layers, and classifiers). 
Note that we exclude the pre-trained feature extractors.
In MulT and TFR-Net, an extra computational complexity of self-attention is introduced for aggregating the repeatedly reinforced representations of each modality, which is not required in MCT-HFR. 
During model training, the feature reconstruction networks can increase the computational cost of TFR-Net, EMT-DLFR, and MCT-HFR. 
Compared with TFR-Net, the trainable parameters and MACs of MCT-HFR drop by 61.5\% and 55.8\%.
Compared with EMT-DLFR, the higher MACs of MCT-HFR are due to the more sophisticated Transformer-based decoders in LFI.
During model inference, MCT-HFR also achieves the lowest computational cost at the model level, with 2$\times$ less trainable parameters and slightly less MACs than the sub-optimal model.

\subsection{Ablation Study} \label{ablation}
To better understand the influence of different modules in our proposed MCT-HFR framework, we perform extensive ablation studies on the IEMOCAP dataset and report the AUILC results for the incomplete testing data by default.
In the following subsections, we employ the ``dynamic incomplete training'' strategy for all models.

\begin{table}[htbp]
\centering
\caption{Ablation results for the re-scaling factors on the incomplete testing data of IEMOCAP dataset with different duration intervals. The scores (\%) of UA are reported, with the best results highlighted in bold.}
\label{tab:ablation-scaling}
\begin{tabular}{lcccccc}
\toprule
\multirow{2}{*}{Method} & \multicolumn{6}{c}{Duration of Testing Data} \\
\cmidrule(lr){2-7}
                       & < 2 s (20.9\%)   & 2\textasciitilde4 s (41.7\%)    & 4\textasciitilde6 s (18.7\%)     & 6\textasciitilde8 s (9.6\%)  & > 8 s (9.1\%) & Overall  \\
\midrule      
PMR      & 61.88  &  64.69 & 66.76 & 71.63 & 72.72 & 66.04       \\
TFR-Net  & 65.09  &  65.51 & 65.92 & 71.70 & 73.18 & 67.08       \\
\midrule 
MCT-HFR                                 & \textbf{65.21}  & \textbf{65.86}  & \textbf{68.39}  & \textbf{74.14}  & \textbf{75.86}  & \textbf{68.06}  \\
\quad w/o $\gamma_{b}$                  & 64.63  & 65.45  & 66.87 & 73.28 & 75.23 & 67.36  \\
\quad w/o $\gamma_{e}$                  & 63.29  & 64.23  & 65.60 & 71.37 & 73.73 & 66.14 \\
\quad w/o $\gamma_{b},\gamma_{e}$       & 62.58  & 63.11  & 66.06 & 72.44 & 70.51 & 65.54 \\
\bottomrule
\end{tabular}
\end{table}

\subsubsection{Effects of the Re-scaling Factors in MCT}
As illustrated in Section~\ref{MCT}, the re-scaling factors $\gamma_{b}$ and $\gamma_{e}$ are introduced in constructing the keys of hyper-modality.
The motivation is to address the issue of modality imbalance and attention dispersion caused by excessively long concatenated sequences.
To fully investigate the effects of the re-scaling factors, we ablate $\gamma_{b}$ and $\gamma_{e}$ respectively, then evaluate the models on the testing data split by different duration intervals.
As shown in Table~\ref{tab:ablation-scaling}, the models' performances generally increase as the duration grows due to more comprehensive contextual information.
However, we can observe a performance decline from ``6\textasciitilde8 s'' to ``>8 s'' for the MCT-HFR model without the re-scaling factors, which implies its insufficient ability to handle long sequences.
By adding either $\gamma_{b}$ or $\gamma_{e}$, the performance of MCT-HFR improves on most duration intervals, especially on the ``>8 s'' one.
Moreover, the combination of these two re-scaling factors can lead to the best performing model, proving the effectiveness of the re-scaling factors in MCT.

\subsubsection{Roles of LFI and GFA in Feature Reconstruction}
In our proposed MCT-HFR framework, the feature reconstruction related branch consists of the Local Feature Imagination (LFI) and Global Feature Alignment (GFA) modules.
To verify the roles of these components, we implement the following three systems for comparison:
\begin{itemize}
 \item \textbf{MCT-GFA:} It comes from MCT-HFR but omits the LFI module, \emph{i.e.}, it reconstructs features only from the global perspective.
 \item \textbf{MCT-LFI:} It comes from MCT-HFR but omits the GFA module, \emph{i.e.}, it reconstructs features only from the local perspective.
 \item \textbf{MCT:} It comes from MCT-HFR but omits the LFI and GFA modules, \emph{i.e.}, it contains only the emotion recognition task related branch.
\end{itemize}

\begin{figure*}[htb]
  \centering
  \scalebox{1.0}
  {\includegraphics[width=\linewidth]{ 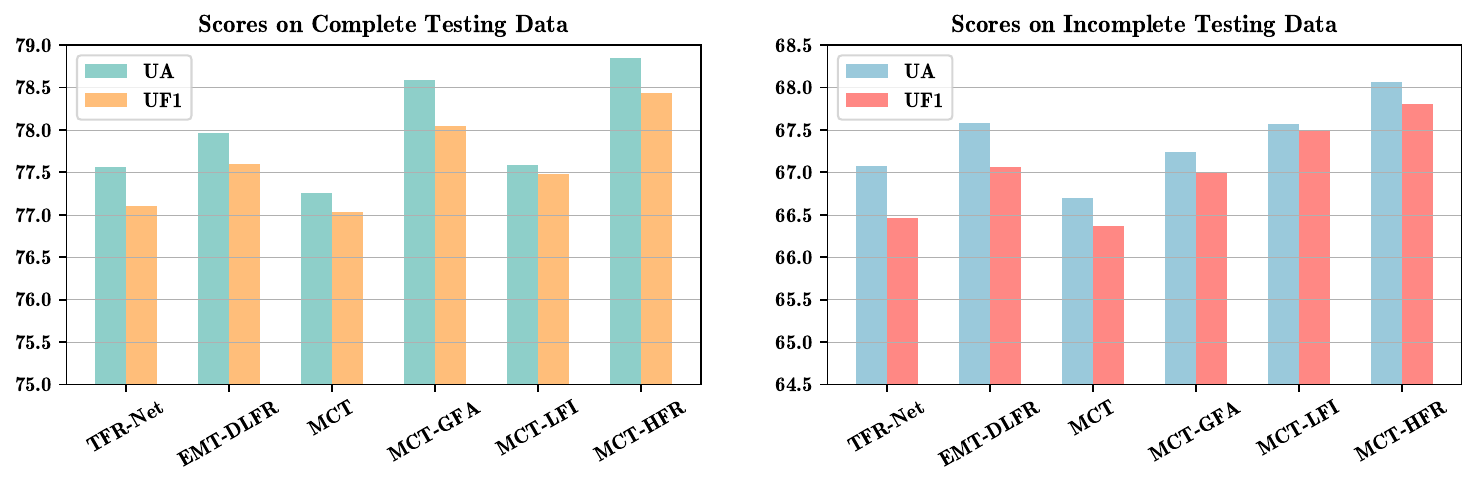}}
  \caption{Ablation results for the components of HFR on the IEMOCAP dataset. The scores (\%) of UA and UF1 are reported for both the complete and incomplete testing data.}
  \label{fig:ablation_loss}
\end{figure*}

The experimental results of different models are shown in Fig.~\ref{fig:ablation_loss}. 
We observe that both MCT-GFA and MCT-LFI exhibit performance gain over MCT, proving that the feature reconstruction from both local and global perspectives can improve the basic recognition ability and model robustness against feature missing.
Specifically, MCT-GFA outperforms MCT-LFI on the complete testing data, while MCT-LFI surpasses MCT-GFA on the incomplete testing data.
We believe that GFA can better capture the global semantics at relatively lower feature missing rates, while LFI can effectively relieve the performance decline as the missing rate grows.
Consequently, MCT-HFR achieves the best performing by combining these two complementary modules together.
For better comparison, we also present the results of TFR-Net and EMT-DLFR in Fig.~\ref{fig:ablation_loss}.
We observe that even MCT can achieve competitive performance with TFR-Net, indicating that the collaborative modality reinforcements of MCT are beneficial to the model robustness.

\subsubsection{Distance Metrics for GFA}
\begin{figure*}[htb]
  \centering
  \scalebox{1.0}
  {\includegraphics[width=\linewidth]{ 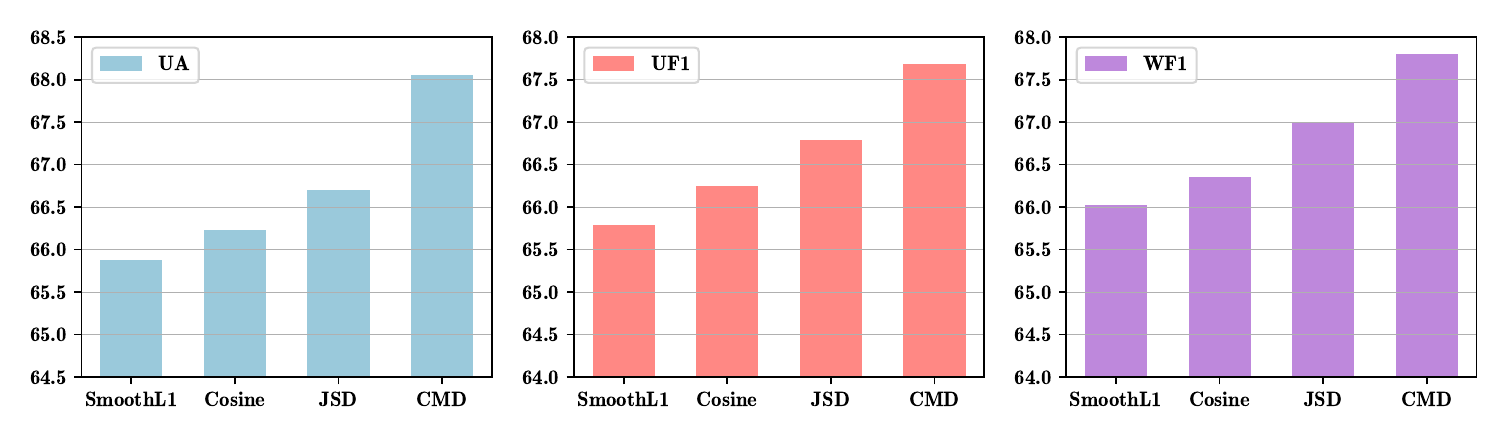}}
  \caption{Ablation results for the distance metrics of GFA on the IEMOCAP dataset. The scores (\%) of UA, UF1 and WF1 are reported for the incomplete testing data.}
  \label{fig:ablation_gfa}
\end{figure*}

As illustrated in Section~\ref{HFR}, we employ CMD as the distance metric in GFA to measure the discrepancy between the global semantics of complete and incomplete views of data.
In this subsection, we compare CMD with another three distance metrics, \emph{i.e.}, smooth L1 loss, cosine similarity, and JSD.
The experimental results are depicted in Fig.~\ref{fig:ablation_gfa}.
Compared with the other three distance metrics, CMD exhibits the great advantage for enhancing the model robustness.
Among the four metrics, smooth L1 loss turns out to be the worst performing metric, mainly due to the excessive constraint on the global semantic spaces.
Besides, we conjecture that the relatively poor performance of cosine similarity can be due to the potential risk of shortcut learning~\cite{DuMJDDGSH21shortcut}.
While JSD is based on mean (first raw moment) matching of statistical moments, CMD can explicitly match the higher order moments of the view-specific distributions. 
In this way, CMD helps to align the global semantics more comprehensively.

\begin{figure*}[htb]
  \centering
  \scalebox{1.0}
  {\includegraphics[width=\linewidth]{ 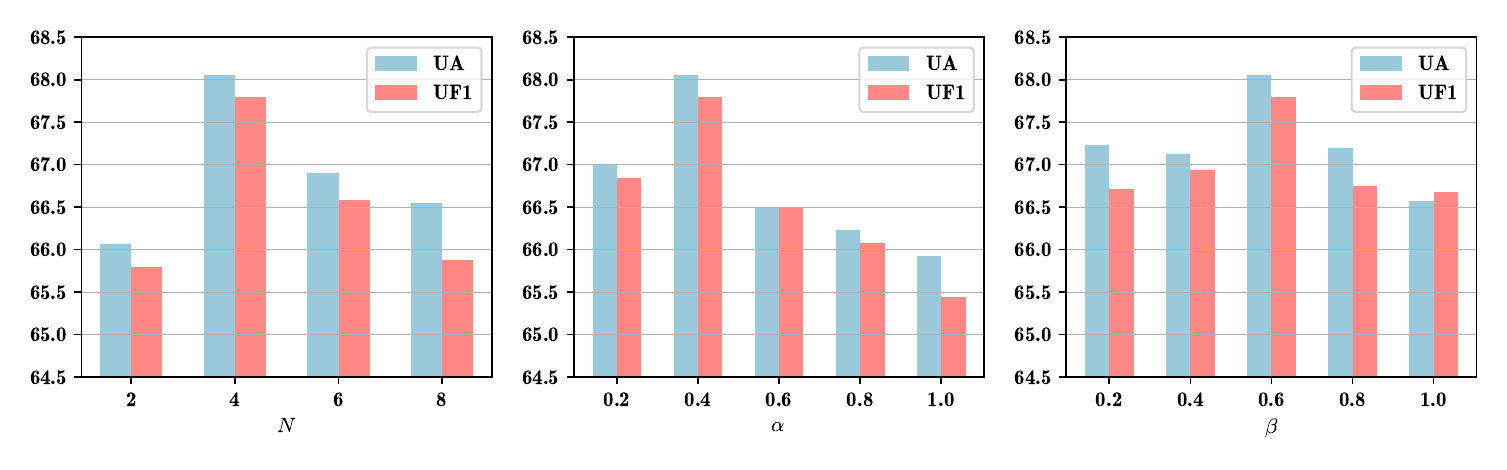}}
  \caption{Ablation results for different hyper-parameters $N$, $\alpha$ and $\beta$ on the IEMOCAP dataset. The scores (\%) of UA and UF1 are reported for the incomplete testing data.}
  \label{fig:ablation_hyperparams}
\end{figure*}

\subsubsection{Hyper-parameter Analysis}
In this part, we investigate the effects of different stacked layers ($N$) in MCT, along with the controlling factors ($\alpha$ and $\beta$) in the objective loss function, respectively.
From the results presented in Fig.~\ref{fig:ablation_hyperparams}, we conclude the following observations:

1) The performance of MCT-HFR significantly improves as $N$ grows from 2 to 4, indicating that the stacked layers can deepen the interactions across modalities.
Nevertheless, MCT-HFR suffers from a performance decline as $N$ continues to increase.
Considering the limited scale of the IEMOCAP dataset, we believe that this may be resulted from the overfitting issue.

2) As illustrated in Section~\ref{implementation}, grid searching is conducted to determine the optimal controlling factors, with $\alpha=0.4$ and $\beta=0.6$.
By fixing one factor and adjusting the other one, we observe that $\alpha$ is more sensitive to the hyper-parameter tuning, with a considerable performance decline when $\alpha$ equals 1.
Considering the relatively larger values of $\mathcal{L}_{\operatorname{GFA}}$, we speculate that the feature reconstruction loss may overwhelm the primary task loss with relatively higher $\alpha$.
Consequently, the overall robustness of MCT-HFR may be improved by carefully balancing the importance of different losses in the objective loss function.

\begin{figure*}[htb]
  \centering
  \scalebox{0.8}
  {\includegraphics[width=\linewidth]{ 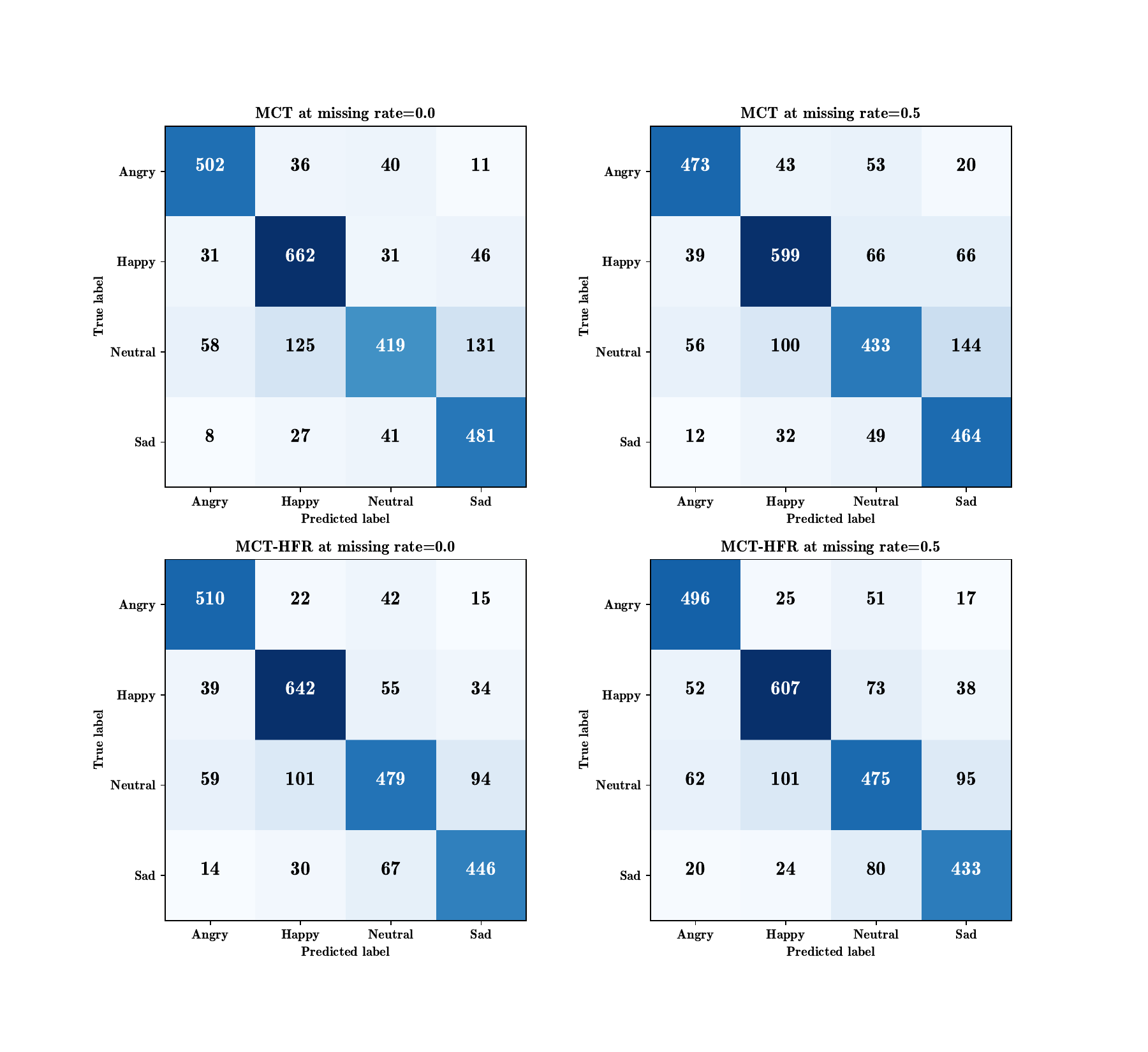}}
  \caption{ Visualization for confusion matrices of MCT and MCT-HFR on the IEMOCAP dataset with feature missing rates of 0.0 and 0.5, respectively.}
  \label{fig:visualize_cm}
\end{figure*}

\subsection{Visualization Analysis}
In this subsection, we perform the following visualization experiments to qualitatively analyze the effectiveness of our proposed method.

\subsubsection{Visualization of Confusion Matrices}
Fig.~\ref{fig:visualize_cm} visualizes the confusion matrices of MCT and MCT-HFR on both complete and incomplete testing set.
We find that the ``happy'', ``angry'' and ``sad'' emotions are most likely to be confused with the ``neutral'' emotion, mainly due to the predominant proportion of ``neutral'' samples on the IEMOCAP dataset.
By integrating HFR into MCT, the recognition accuracy of the ``neutral'' emotion improves from 57.16\% to 65.35\% on the complete testing data.
Moreover, HFR considerably relieve the accuracy decline of all the emotion classes as the feature missing rate increases from 0.0 to 0.5.
As a result, MCT-HFR outperforms MCT in both complete and incomplete data scenarios.

\subsubsection{Visualization of Embedding Spaces}
\begin{figure*}[htb]
  \centering
  \scalebox{1.0}
  {\includegraphics[width=\linewidth]{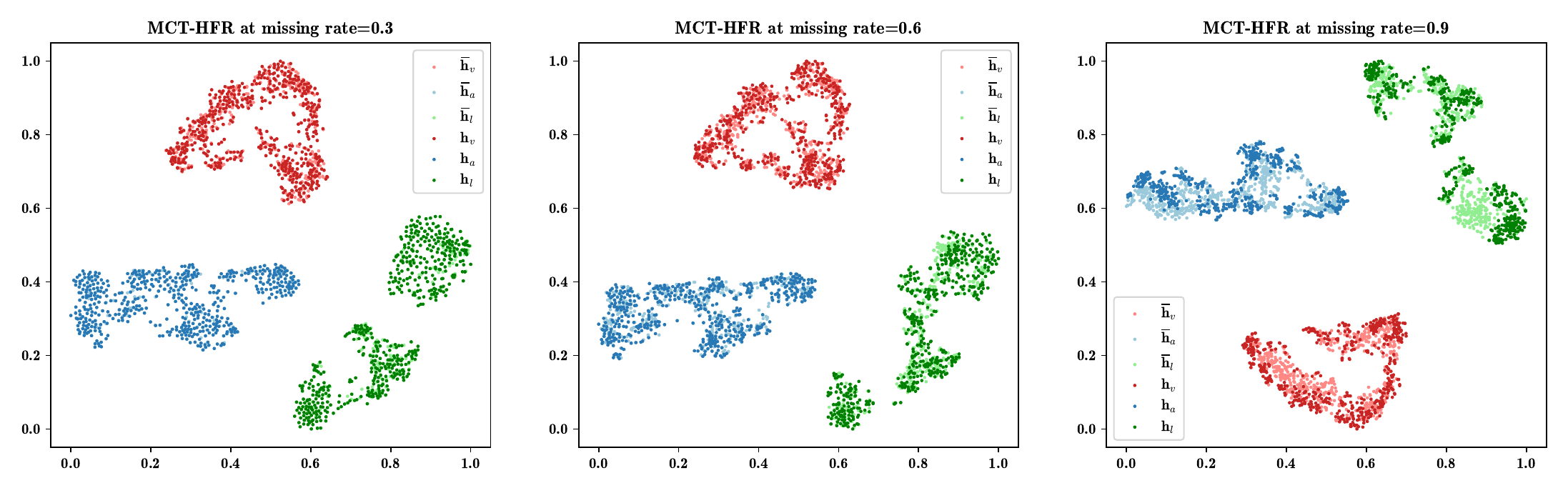}}
  \caption{T-SNE visualization for the modality-wise latent representations of MCT-HFR from complete and incomplete views.
  The results are derived from the first-fold testing set of the IEMOCAP dataset with feature missing rates of 0.3, 0.6 and 0.9, respectively.}
  \label{fig:visualize_ebd}
\end{figure*}

Fig.~\ref{fig:visualize_ebd} visualizes the distribution of the multimodal embeddings from the complete and incomplete views of testing data using t-SNE~\cite{JMLRvandermaaten08a/tsne}.
We observe that the distributions of the ``complete'' and ``incomplete''  embeddings are very close at a relatively lower feature missing rate.
As the missing rate grows, the deviations of specific distribution positions become more apparent. 
Nonetheless, the boundaries of overall distributions in the same modality are still very similar, indicating the superior feature reconstruction ability of MCT-HFR even in severely feature missing cases.

\section{Conclusions} \label{conclusion}
In this paper, we present a unified framework, MCT-HFR, to promote the efficiency and robustness of multimodal emotion recognition in real-world scenarios.
Instead of stacking vanilla attention units, MCT is specifically optimized for unaligned multimodal input, with the key innovation of modality-wise parameter sharing and parameter-free re-scaling mechanism. 
By collaborative modality reinforcements, MCT can surpass the most advanced Transformer-based models with noticeably less space and time complexity.
During model training, HFR is introduced as an auxiliary supervision to further enhance the robustness of MCT from both the local and global perspectives.
Experimental results on two benchmark datasets demonstrate the effectiveness of our method.
Through quantitative and qualitative analysis, we first confirm that MCT-HFR consistently outperforms currently advanced approaches with various feature missing rates, achieving the best performance in both complete and incomplete data scenarios.
Later, we perform extensive ablation studies to verify the roles of different modules in this framework.
The visualization analysis further reveals the excellent feature reconstruction ability of MCT-HFR, indicating its superior robustness against random feature missing.

Future work will explore the effectiveness of our method in more generic practical applications, where various data corruptions can occur at the raw input.
For example, we can imitate the low signal-to-noise ratio scenario for raw audio input, and generate the associate transcriptions leveraging automatic speech recognition models.
In addition, we plan to promote the robustness of our method in the cases of entire modality absence as well.

\section*{List of Abbreviations}
\begin{acronym}
  \acro{MER}{Multimodal Emotion Recognition}
  \acro{RMFM}{Random Modality Feature Missing}
  \acro{SSL}{Self-supervised learning}  
  \acro{SAU}{Self-attention Unit}
  \acro{CAU}{Cross-modal Attention Unit}
  \acro{MRAU}{Multimodal Re-scaled Attention Unit}
  \acro{MCT}{Modality-Collaborative Transformer}
  \acro{GFA}{Global Feature Alignment}
  \acro{LFI}{Local Feature Imagination}
  \acro{HFR}{Hybrid Feature Reconstruction}
  \acro{MCT-HFR}{Modality-Collaborative Transformer with Hybrid Feature Reconstruction}
  \acro{CMD}{Central Moment Discrepancy}
  \acro{JSD}{Jensen-Shannon Divergence}
\end{acronym}

\begin{acks}
This research was partially supported by the National Key Research and Development Program of China (No. 2021YFC3320103).
\end{acks}

\bibliographystyle{ACM-Reference-Format}
\bibliography{sample-base}

\end{document}